\documentclass[10pt,twocolumn,letterpaper]{article}

\usepackage{graphicx}
\usepackage{amsmath}
\usepackage{amssymb}
\usepackage{booktabs}
\usepackage{tabularx}
\usepackage[pagebackref,breaklinks,colorlinks]{hyperref}
\usepackage[capitalize]{cleveref}
\crefname{section}{Sec.}{Secs.}
\Crefname{section}{Section}{Sections}
\Crefname{table}{Table}{Tables}
\crefname{table}{Tab.}{Tabs.}

\usepackage{multirow}
\usepackage{svg} 
\usepackage{booktabs} 

\begin{document}

\title{Reimagining Image Segmentation using Active Contour: From Chan Vese Algorithm into a Proposal Novel Functional Loss Framework}

\author{
Gianluca Guzzetta\\
Politecnico di Torino\\
{\tt\small gianluca.guzzetta@studenti.polito.it}
\\
}

\maketitle

\begin{abstract}
   In this paper, we present a comprehensive study and analysis of the Chan Vese algorithm for image segmentation, using a discretized scheme obtained from the empirical study of Chan-Vese model's functional energy and its partial differential equation based on its level set function. The proof to such result is shown, and an implementate using \textit{MATLAB}. Taking advantange of nowadays methodologies in computer vision, we also propose a \textit{functional segmentation loss} based on active contour based on \texttt{pytorch.nn.ModuleLoss} and level set based on Chan-Vese algorithm proposed, and comparing such results to common computer vision segmentation datasets and we compare classical loss performances with the proposed one. All code and and used material are available at \href{https://github.com/gguzzy/cv_functional_loss}{cv\_functional\_loss repository}. 

   
\end{abstract}

\section{Problem overview}
Image segmentation is the process of partitioning a digital image into multiple segments (sets of pixels). Hence, image segmentation consists into portioning a digital image into valuable and reliable pieces of information, easier to analyze. The whole picture does not contain only relevant data, in most computer vision scenarios, we are interested in a portion of area of pixel of the image, which could be represented as people or some certain objects we are interested in, as our cat who crossed the perimeter in the garden. Nowadays the aims are immeasurable. 

Although it is a common problem, image segmentation has a wide set of resolution methods which aim at collecting set of pixels or a set of relevant pixels called objects.

These methods are in general (i.e. not limited to):

\begin{itemize}
    \item \textbf{Thresholding}: This method involves partitioning an image based on pixel intensity. Pixels with intensity above a certain threshold are classified differently from those below, allowing for straightforward segmentation of images into foreground and background.
    
    \item \textbf{Clustering}: This technique partitions image pixels into clusters based on attributes such as color and intensity. K-means clustering is a notable example, categorizing pixels into clusters with similar values.
    
    \item \textbf{Contour-based methods}: These methods aim to find contours within an image, identifying boundaries between different image regions. The active contour model, or snakes, dynamically adjusts contours to outline object edges. 
    
    \item \textbf{Region-based methods}: These methods segment an image based on the homogeneity of pixel regions. Techniques like Region Growing start from seed points and aggregate neighboring pixels with similar properties to form a segment. 
\end{itemize}

In this proposed work, we used an active contour modeling using the Chan-Vese method, an active contour modeling method renowned for its efficacy in capturing complex object boundaries and independent on image intensities. 

\section{Proposed approach}
\subsection{Preprocessing}
In our experiment, an image goes through a well-defined process of image preprocessing, moreover the experiments are run onto different scenarios to prove its effectiveness, hence the starting image is segmented using the proposed method and compared to the noisy images processed through the next proposed approeaches. Indeed noise is introduced in order to add a non linear factor to the problem, as in real world applications.
Each image has been treated, introducing noise and removing those factors later, using the following techniques:

\subsection{Data Augmentation: noise introduction to the dataset}

Data augmentation plays a critical role in the development of machine learning models, particularly in the field of computer vision. By introducing variations in the training data, models can learn to generalize better and become more robust to unseen data. In this work, we focus on the incorporation of Gaussian noise and Salt-and-Pepper (S\&P) noise as augmentation strategies to simulate real-world imperfections in images.

\textbf{Gaussian Noise Addition}: Gaussian noise, characterized by its normal distribution, is added to images to mimic the effect of random variations in pixel values. This is mathematically represented as:
\begin{equation}
I_{\text{noisy}} = I + \mathcal{N}(\mu, \sigma^2),
\end{equation}
where \(I\) is the original image, \(\mathcal{N}(\mu, \sigma^2)\) denotes the Gaussian distribution with mean \(\mu\) and variance \(\sigma^2\), and \(I_{\text{noisy}}\) is the resultant image. This technique aids in training models that are less sensitive to slight pixel intensity variations.

\textbf{Salt-and-Pepper Noise Injection}: S\&P noise, also known as impulse noise, introduces sharp, sudden disturbances in the image, represented by randomly scattering white and black pixels. This type of noise challenges the model's ability to maintain performance despite the presence of significant pixel-level anomalies.

All these techniques aim at improving the model's robustness, meaning the efficiency of the segmentation to low quality datasets. 

\subsection{Feature Engineering: Advanced Noise Removal Techniques}

Effective feature engineering is paramount in preparing data to increase its data quality for our model, especially when dealing with noisy datasets. To counteract the noise introduced during data augmentation, we employ sophisticated noise removal techniques as part of our preprocessing pipeline.

\begin{itemize}
    \item \textbf{Gaussian Noise Elimination via Gaussian Filtering}: Gaussian filtering is a smoothing technique that reduces Gaussian noise. It utilizes a Gaussian kernel to blur the image, effectively diminishing the impact of noise:
    \begin{equation}
    G(x, y) = \frac{1}{2\pi\sigma^2}e^{-\frac{x^2 + y^2}{2\sigma^2}},
    \end{equation}
    where \(G(x, y)\) is the Gaussian kernel, and \(\sigma\) is the standard deviation of the distribution. This method preserves edge properties while reducing noise, making it ideal for preprocessing in image segmentation tasks.
    
    \item \textbf{Salt-and-Pepper Noise Reduction via Median Filtering}: Median filtering is a non-linear process used to remove S\&P noise. It replaces each pixel's value with the median value of the intensities in its neighborhood. This approach effectively preserves edges while removing noise, essential for maintaining the structural integrity of objects in the image.
\end{itemize}



\begin{table}[htbp]
\caption{Filtered images: MSE and RMSE}
\centering
\begin{tabular}{lcc}
\toprule
\textbf{Method} & \textbf{MSE} & \textbf{RMSE} \\
\midrule
Gaussian Filtered & 0.029759 & 0.17251 \\
Median Filtered & 0.00061407 & 0.02478 \\
Gaussian S\&P Filtered & 0.0030896 & 0.055584 \\
Median S\&P Filtered & 0.0030237 & 0.054988 \\
Median Gaussian S\&P Filtered & 0.0030237 & 0.054988 \\
\bottomrule
\end{tabular}
\label{tab:MSE_RMSE}
\end{table}

As evidenced in our final benchmark tests in Figure \ref{fig:comparisonPreprocessing}, even with the introduction of random noise, the models maintain high performance levels. This underscores the effectiveness of our proposed augmentation and preprocessing strategies, demonstrating the model's robustness against various noise types.

\begin{figure}[htbp]
  \centering
  \includegraphics[width=3.33in]{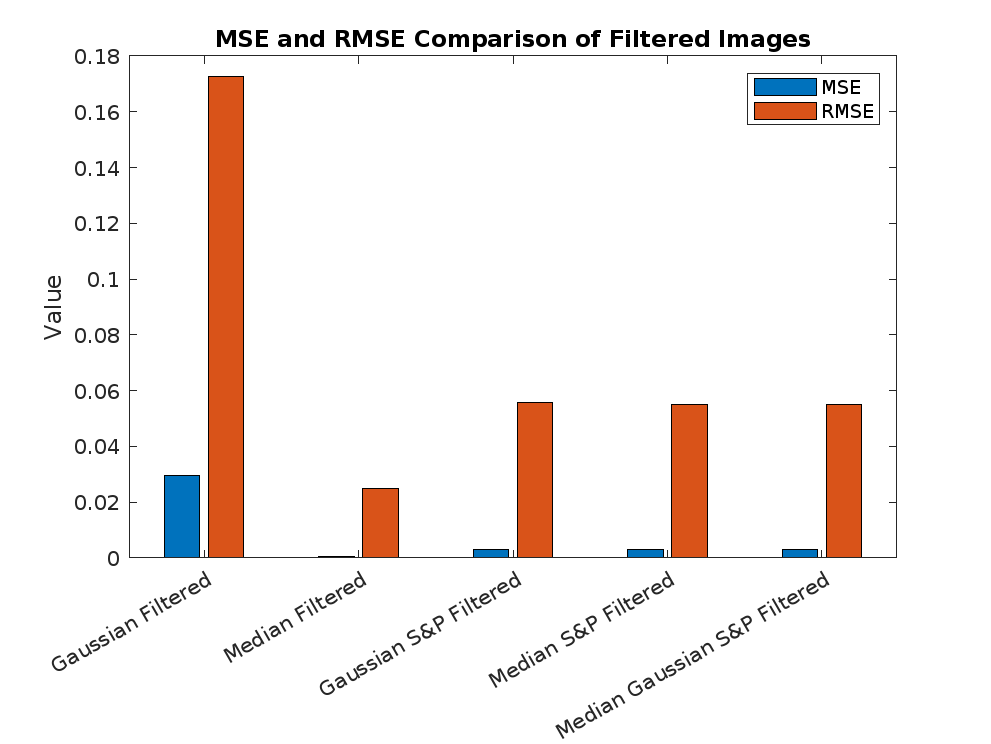}
  \label{fig:comparisonPreprocessing}
\end{figure}

\subsection*{From Mumford-Shah to Chan-Vese}
Introduced by Mumford and Shah in their seminal 1989 paper \cite{mumford1989optimal}, the Mumford-Shah model has been a cornerstone in the field of image segmentation. The model seeks to partition an image into regions of approximately uniform intensity by minimizing an energy functional.
For a given image $I$ defined over a domain $\Omega \in \mathbb{R}^2$, the Mumford-Shah functional can be expressed as:

\subsection*{The Mumford-Shah model}

\begin{align}
E(u, C) = & \int_{\Omega \setminus C} |\nabla u|^2 \, dx \nonumber \\
& + \mu \int_{\Omega \setminus C} (u - I)^2 \, dx \nonumber \\
& + \nu \, \text{Length}(C),
\label{eq:shah-energy}
\end{align}

where:
\begin{itemize}
    \item $u$ is an approximation of the image
    \item $I$ within the domain $\Omega \setminus C$
    \item $C$ denotes the set of edges in the image
    \item $\mu$, $\nu$ are regularization parameters, chosen arbitrarily by the user. 
\end{itemize}

In the Mumford-Shah functional, \( \Omega \) represents the entire image domain, while \( \Omega \setminus C \) denotes the image domain excluding the set of edges \( C \), essentially covering the regions within the image that are not part of the segmentation boundaries. 
The parameter \( \mu \) controls the trade-off between the image's fidelity to its approximation \( u \) and smoothness, whereas \( \nu \) regulates the importance of the contour's length in the overall segmentation, aiming to minimize unnecessary segmentation complexity.
The functional aims to balance the smoothness of $u$ and the fidelity to the original image $I$, along with the complexity of the segmentation represented by $\text{Length}(C)$.

The model aims to find piecewise smooth representations of images by minimizing an energy function that penalizes the image's deviation from the model and the length of the contours.

Expanding upon the Mumford-Shah model, Chan and Vese introduced a simplified version in 2001 \cite{chan2001active}, targeting images with two distinct intensity levels. This model, known as the Chan-Vese model, is designed for binary segmentation tasks, optimizing a similar energy functional to directly distinguish between the two regions \cite{chan2001active}. 

While the original Mumford-Shah model does not inherently rely on level set functions, its adaptations, including the Chan-Vese model, often utilize level set methods to represent the evolving curves in the segmentation process. This approach allows for a more flexible representation of contours and simplifies the numerical implementation of the segmentation task.

\subsection*{The Chan-Vese model}
The Chan-Vese model is a specialization of the Mumford-Shah model for segmenting images into two distinct regions. It uses an energy function defined as:

\begin{align}
F(c_1, c_2, C) = & \, \mu \, \text{Length}(C) \nonumber \\
& + \, \nu \, \text{Area}(\text{int}(C)) \nonumber \\
& + \, \lambda_1 \int_{\text{int}(C)} |u_0 - c_1|^2 \, dx \nonumber \\
& + \, \lambda_2 \int_{\text{ext}(C)} |u_0 - c_2|^2 \, dx
\label{eq:chan-vese-energy}
\end{align}

where $c_1$ and $c_2$ represent the mean intensities inside and outside the contour $C$. The terms $\lambda_1$, $\lambda_2$, and $\mu$ are weighting parameters that control the segmentation's adherence to the model's assumptions. 

This model effectively segments images by evolving a contour $C$ to minimize the above energy, adapting well to images with homogenous regions.

The Chan-Vese model simplifies this approach by assuming images with two intensity levels, optimizing a similar energy function to distinguish between these regions directly.

\textbf{Example}: The Chan-Vese algorithm excels in segmenting images with clear object boundaries against a uniform background, efficiently handling variations in intensity and texture without prior knowledge of object location.

\subsection{Level Set Methods}

Level set methods play a pivotal role in the domain of computational image segmentation, offering a robust framework for dynamically evolving contours. Within the Chan-Vese model, these methods facilitate the optimization of an energy functional that guides the segmentation process. The level set function, \(\phi(x,y,t)\), encapsulates the contour \(C\) implicitly as its zero level set, evolving under a carefully designed partial differential equation (PDE). This evolution is meticulously driven to minimize the energy functional, as detailed in Equation \ref{eq:chan-vese-energy}.

where \(c_1\) and \(c_2\) denote the mean intensities within and outside the evolving contour \(C\), respectively. The weighting parameters \(\lambda_1\), \(\lambda_2\), and \(\mu\) are calibrated to balance the trade-offs inherent in the segmentation task, ensuring an adherence to the image's inherent structures while promoting contour smoothness and minimal complexity.

\subsubsection{The Lipschitz Function}

A critical aspect of the level set formulation is the initialization of the level set function \(\phi_0(x,y)\), which should ideally be a Lipschitz function. This property is crucial for ensuring the numerical stability and convergence of the evolving contour. A Lipschitz function adheres to the condition:

\begin{equation}
|f(x) - f(y)| \leq L \|x - y\|,
\end{equation}

for every pair \(x, y\) within the domain, where \(L\) represents the Lipschitz constant. This constraint ensures a controlled evolution pace, preventing excessive, abrupt changes in the contour's geometry.

\textbf{Initialization with a Circle or Ellipse}: For initializing the level set function in the Chan-Vese framework, employing geometric shapes such as circles or ellipses offers a pragmatic and theoretically sound approach. Both configurations qualify as Lipschitz functions, attributable to their smooth, continuous boundaries that can be mathematically articulated with bounded derivatives.

A circular initial contour centered at \((x_0, y_0)\) with radius \(r\) is defined by:

\begin{equation}
\phi_0(x,y) = \sqrt{(x - x_0)^2 + (y - y_0)^2} - r,
\end{equation}

demonstrating Lipschitz continuity as the magnitude of its gradient is uniformly bounded \cite{bregman1967relaxation}. Similarly, an elliptical contour, characterized by its semi-major and semi-minor axes \(a\) and \(b\), respectively, the subtraction by 1 adjusts the level set to position the zero level set contour exactly at the boundary of the ellipse. This is the correct formulation for representing an elliptical contour in terms of a level set function; it is represented as:

\begin{equation}
\phi_0(x,y) = \sqrt{\left(\frac{x - x_0}{a}\right)^2 + \left(\frac{y - y_0}{b}\right)^2} - 1,
\end{equation}

which also exhibits bounded gradients, affirming its suitability as an initial Lipschitz condition for the level set evolution. 

The employment of these geometric forms as initial conditions not only simplifies the numerical implementation but also aligns with the theoretical underpinnings required for a stable and efficient segmentation process. This approach underscores the Chan-Vese model's versatility in adapting to images with diverse intensity profiles, facilitating precise segmentation outcomes.

\section*{Level Set Formulation}
Instead of searching for the solution in terms of \( C \), we can redefine the problem in the level set formalism. In the level set method, \( C \subseteq \Omega \) is represented by the zero level set of some Lipschitz function \( \Phi: \Omega \rightarrow \mathbb{R} \), such that:
\begin{equation}
\begin{cases}
    C = \{ (x,y) \in \Omega : \Phi(x,y) = 0 \} \\
    \text{inside}(C) = \{ (x,y) \in \Omega : \Phi(x,y) > 0 \} \\
    \text{outside}(C) = \Omega \backslash \text{inside}(C) = \{ (x,y) \in \Omega : \Phi(x,y) < 0 \}
\end{cases}
\end{equation}

\subsection*{Energy Functional}
The energy functional in terms of \( \Phi \) is given by:
\begin{align}
F(c_1, c_2, \Phi) = & \, \mu \int_{\Omega} \delta(\Phi(x,y))\|\nabla \Phi(x,y)\|\, dx\,dy \nonumber \\
& + \nu \int_{\Omega} H(\Phi(x,y))\, dx\,dy \nonumber \\
& + \lambda_1 \int_{\Omega} |u_0(x,y) - c_1|^2 H(\Phi(x,y))\, dx\,dy \nonumber \\
& + \lambda_2 \int_{\Omega} |u_0(x,y) - c_2|^2 (1 - H(\Phi(x,y)))\, dx\,dy
\label{eq:centered-updated-chan-vese-energy}
\end{align}

where:
\begin{itemize}
    \item \( \mu \) is the parameter that weights the length of the contour, controlling the smoothness of the segmentation boundary; 
    \item \( \nu \) weights the area inside the contour, affecting how the area influences the segmentation; 
    \item \( \lambda_1 \) and \( \lambda_2 \) weight the intensity differences within and outside the contour, respectively, aiding in fitting the model to the image data;
    \item  \( H \) is the Heaviside function, used to differentiate the regions inside and outside the contour; 
    \item \( x \) and \( y \) are the spatial coordinates in the image domain \( \Omega \); 
    \item \( \nabla \Phi \) is the gradient of \( \Phi \), with its norm representing the rate of change of \( \Phi \), related to the curvature of the contour \( C \) in the level set method.
    \item  The domain \( \Omega \) represents the area over which the image is defined and the segmentation is performed.
\end{itemize}

\subsection*{Partial Differential Equation for \( \Phi \)}
The function \( \Phi \) which minimizes the energy functional satisfies the PDE:
\begin{align}
\frac{\partial \Phi}{\partial t} = \delta(\Phi) \bigg( & \mu \text{div} \left( \frac{\nabla \Phi}{\|\nabla \Phi\|} \right) - \nu \nonumber \\
& - \lambda_1(u_0 - c_1)^2 + \lambda_2(u_0 - c_2)^2 \bigg)
\label{eq:level-set-evolution}
\end{align}

The curvature \( \kappa(\Phi) \) \cite{aubert2003image} of the evolving contour is given by the spatial derivatives of \( \Phi \) up to the second order:
\begin{equation}
\kappa(\Phi) = \frac{\Phi_{xx}\Phi_{y}^2 - 2\Phi_{xy}\Phi_{x}\Phi_{y} + \Phi_{yy}\Phi_{x}^2}{(\Phi_{x}^2 + \Phi_{y}^2)^{3/2}}
\end{equation}

\subsection*{Numerical Scheme}

First, we introduce the regularization of the Heaviside and Dirac delta functions for some constant $\epsilon > 0$ as follows:
\begin{align*}
    H_\epsilon(x) &= \frac{1}{2} \left(1 + \frac{2}{\pi} \arctan\left(\frac{x}{\epsilon}\right)\right), \\
    \delta_\epsilon(x) &= \frac{1}{\pi}\frac{\epsilon}{\epsilon^2 + x^2}.
\end{align*}

These regularizations are used in simulations to lead to the global minimum of the energy efficiently. Choosing $\epsilon = h$, where $h$ is the space step, is reasonable since $h$ is the smallest space step in the problem. In the available case we used  $\epsilon = h = 1$.

\subsection*{Discretization of the PDE}

Let $\Phi^n_{i,j} = \Phi(n\Delta t, x_i, y_j)$, where $\Delta t$ is the time step. The PDE can be discretized using spatial finite differences as follows:
\\
\begin{align*}
    \Delta_x \Phi^n_{i,j} &= \frac{\Phi^n_{i+1,j} - \Phi^n_{i,j}}{h}, & \Delta_x^- \Phi^n_{i,j} &= \frac{\Phi^n_{i,j} - \Phi^n_{i-1,j}}{h}, \\
    \Delta_y \Phi^n_{i,j} &= \frac{\Phi^n_{i,j+1} - \Phi^n_{i,j}}{h}, & \Delta_y^- \Phi^n_{i,j} &= \frac{\Phi^n_{i,j} - \Phi^n_{i,j-1}}{h}.
\end{align*}

The linearized discretized PDE becomes:
\begin{align*}
    \frac{\Phi^{n+1}_{i,j} - \Phi^n_{i,j}}{\Delta t} = & \, \delta_h(\Phi^n_{i,j}) \Bigg[ \mu \Bigg( \frac{\Delta_x^+ \Phi^n_{i,j} + \Delta_x^- \Phi^n_{i,j}}{2h^2} \\
    & + \frac{\Delta_y^+ \Phi^n_{i,j} + \Delta_y^- \Phi^n_{i,j}}{2h^2} \Bigg) \\
    & - \delta_h(\Phi^n_{i,j}) \Big(v\lambda + \frac{1}{u_c^{i,j}} - \frac{1}{2\Phi^n_{i,j} - \lambda_{u_c^{i,j}}} \Big) \Bigg].
\end{align*}

\subsubsection*{Constants Definition}

Defining constants for a given $\Phi^n$ as: \\
\begin{align*}
    C_1 &= \frac{1}{4} \left( \Phi^n_{i,j+1} - \Phi^n_{i,j-1} \right)^2, \\
    C_2 &= \frac{1}{4} \left( \Phi^n_{i+1,j} - \Phi^n_{i-1,j} \right)^2, \\
    C_3 &= \frac{1}{4} \left( \Phi^n_{i,j+1} - \Phi^n_{i,j-1} \right)^2, \\
    C_4 &= \frac{1}{4} \left( \Phi^n_{i+1,j} - \Phi^n_{i-1,j} \right)^2.
\end{align*}
\\

The simplified equation then is:
\begin{align*}
    \Phi^{n+1}_{i,j} = \Phi^n_{i,j} + \Delta t \delta_h(\Phi^n_{i,j}) \mu \left( C_1 + C_2 + C_3 + C_4 \right).
\end{align*}

\subsection*{Summary of the Algorithm}

\begin{enumerate}
    \item Initialize $\Phi^0_{i,j}$ to some Lipschitz function $\Phi_0$.
    \item Compute $c_1$ and $c_2$ using the regularized Heaviside function.
    \item Solve the PDE using the discretized equation.
    \item Reinitialize $\Phi^{n+1}_{i,j}$ to be the signed distance function to its zero level set using the reinitialization equation.
    \item Check whether the solution is stationary. If not, continue; otherwise, stop.
\end{enumerate}

The process of evolving the level set function $\Phi$ during image segmentation aims to minimize an energy functional that accurately represents the object's boundary within the image. This iterative update of $\Phi$ continues until it converges to a stationary state, which signifies that the contour of the object has been precisely delineated.

The update mechanism for $\Phi$ at each grid point $(i, j)$ from iteration $n$ to $n+1$ relies on the image data and the curvature of $\Phi$ itself. This iterative update is terminated when the changes in $\Phi$ between successive iterations become insignificant across the entire image domain, implying that $\Phi$ has reached a stationary state and further iterations would not meaningfully alter its configuration. The mathematical expression for this stopping criteria is as follows:

The iterative process ceases when the discrepancy between $\Phi^{n+1}_{i,j}$ and $\Phi^n_{i,j}$ is lesser than $\Delta t \cdot h^2$, indicating the attainment of a stationary state. This condition is verified through the computation of a quantity $Q$, which assesses the stationarity of $\Phi$ across iterations:

\[
Q = \sum \left( \left| \Phi^{n+1}_{i,j} - \Phi^n_{i,j} \right| < \Delta t \cdot h^2 \right)
\]

In this context, $\Delta t$ symbolizes the timestep size, and $h$ signifies the spatial resolution of the grid. The summation traverses all grid points $(i,j)$, with the inner comparison producing a Boolean value that denotes whether the change at each point is beneath the predefined threshold. The algorithm concludes when a significant majority of the points meet this stipulation, suggesting that $\Phi$ has become stationary.

For the purpose of reinitializing $\Phi$, the ensuing equation is utilized to guarantee the maintenance of the signed distance function property, which is pivotal for the numerical stability and precision of the level set method:

\[
\Phi^{n+1}_{i,j} = \Phi^n_{i,j} - \Delta t \cdot \text{sign}(\Phi(x,y,t)) \cdot G(\Phi^n_{i,j})
\]

Herein, $\text{sign}(\Phi(x,y,t))$ ensures the directionality of the update respects the gradient of the level set function, aiming to minimize deviations from a signed distance function. Concurrently, $G(\Phi^n_{i,j})$ is a function derived from the gradient of $\Phi$, adjusted to realign its values towards sustaining the signed distance property.

\subsection{Chan-Vese Loss Implementation for RGB Images in PyTorch}
In the pursuit of enhancing image segmentation models, we introduce the \texttt{ChanVeseLossPyTorchRGB}, a PyTorch module designed to adapt the classical Chan-Vese loss for RGB images. This module is particularly tailored for segmentation tasks, leveraging the characteristics of RGB images to compute the loss more effectively.

\subsubsection{Implementation Details}
The PyTorch module \texttt{ChanVeseLossPyTorchRGB} implements this adapted loss function. It requires the predicted segmentation masks, the ground truth masks, and the original RGB images as inputs. The intensity difference loss, $\mathcal{L}_{\text{intensity}, c}$, is computed for each RGB channel and averaged. The smoothness term, $\mathcal{L}_{\text{smoothness}}$, utilizes the gradient of the predicted mask, emphasizing the regularization of the segmentation boundary.

The key components of this module are:

\begin{itemize}
    \item The \texttt{heaviside} function, approximated using the hyperbolic tangent to ensure a smooth transition, facilitating the differentiation process required for gradient descent optimization.
    \item The \texttt{mean\_intensities} function, which calculates the average intensities inside and outside the segmented region for each channel, crucial for the intensity difference computation.
    \item The computation of the smoothness loss, employing the gradient of the sigmoid-activated predictions, to enforce boundary smoothness.
\end{itemize}

The incorporation of this loss function into segmentation models promises enhanced performance by effectively leveraging the color information inherent in RGB images and enforcing smooth, coherent segmentations.

\texttt{Python} code snippet for the \texttt{ChanVeseLossPyTorchRGB} module:


See \href{https://github.com/gguzzy/cv_functional_loss}{code here}.

\section{Results}
A solution in \texttt{MATLAB} code is provided at the end for initialization of level set functions, the evolution of the curvature using the above discretized method, able to cope the dinamic solution of numerical systems.


Consider some assumptions have been done, as for instance using a circle as initial level set method for final test, since it is common use in the literature (look for reference and citation), moreovfer $dy=dx=1$ has been consider to 1 supposing that images are equally spaced, $epsilon=h=1$ since we are dealing with square images, each step should be of length 1 (moving of one step at the time). At last, the area is set to 0 since it doesnt efficiently help with the image segmentation resolution.

Here below there are few experimental results from the aforementioned resolution method:

\subsection{Grey images}

\begin{figure}[htbp]
  \centering
  \includegraphics[width=0.48\linewidth]{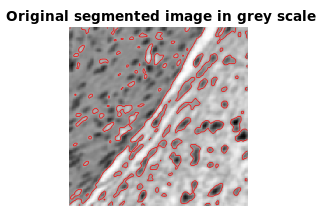}
  \caption{Original segmented image in grey scale}
  \label{fig:immagine_segmentata_con_mask_grey}
\end{figure}

\begin{figure}[htbp]
  \centering
  \includegraphics[width=0.35\linewidth]{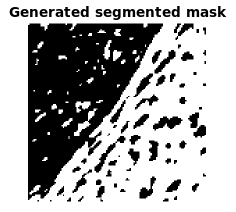}
  \caption{Segmented image result in grey scale}
  \label{fig:segmentation_mask_grey}
\end{figure}

\subsection{RGB images}
When processing RGB images, as in \ref{fig:rgbImages} with the segmentation algorithm, it is imperative to consider the distinct nature of each RGB channel. Consequently, the algorithm is executed separately on each of the RGB channels, resulting in three preliminary segmentation maps. To synthesize these into a final, comprehensive segmentation result, an aggregation step is employed, utilizing logical operations such as AND, OR, and Majority Vote. The aggregation process is delineated as follows:

\begin{figure}[htbp]
  \centering
  \includegraphics[width=0.35\linewidth]{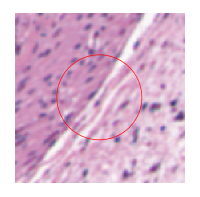}
  \caption{Original image with initial level set function to circle}
  \label{fig:comparisonPreprocessing}
\end{figure}


\begin{itemize}
    \item \textbf{Logical AND}: This operation aggregates the segmentation maps by considering a pixel in the final segmentation to be part of the object if and only if it is identified as part of the object in all three RGB channel segmentations. Mathematically, for a given pixel position $(i, j)$, the final segmentation $S_{i,j}$ is defined as:
    \[
    S_{i,j} = R_{i,j} \land G_{i,j} \land B_{i,j}
    \]
    where $R_{i,j}$, $G_{i,j}$, and $B_{i,j}$ represent the segmentation decisions at pixel $(i, j)$ for the Red, Green, and Blue channels, respectively.
    
    \item \textbf{Logical OR}: In contrast, the OR operation aggregates the maps by marking a pixel as part of the object if it is identified as such in any one of the RGB channel segmentations. Thus, for each pixel $(i, j)$:
    \[
    S_{i,j} = R_{i,j} \lor G_{i,j} \lor B_{i,j}
    \]
    
    \item \textbf{Majority Vote}: This approach determines a pixel's inclusion in the final segmentation based on a majority rule among the three channels. A pixel is considered part of the object if at least two out of the three channel segmentations agree on its inclusion. Formally:
    \[
    S_{i,j} = \text{Majority}(R_{i,j}, G_{i,j}, B_{i,j})
    \]
    where $\text{Majority}(R_{i,j}, G_{i,j}, B_{i,j})$ returns the majority value among $R_{i,j}$, $G_{i,j}$, and $B_{i,j}$.
\end{itemize}

Following the segmentation of each RGB image channel and the subsequent aggregation via these logical operations, the resulting segmentation maps are illustrated below each corresponding RGB image, showcasing the impact of the chosen aggregation method on the final segmentation outcome.


\begin{figure}[htbp]
  \centering
  \includegraphics[width=0.5\linewidth]{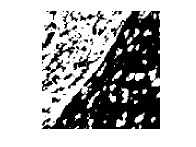}
  \caption{Generated mask after segmentation on original image: channel R}
  \label{fig:comparisonPreprocessing}
\end{figure}

\begin{figure}[htbp]
  \centering
  \includegraphics[width=0.5\linewidth]{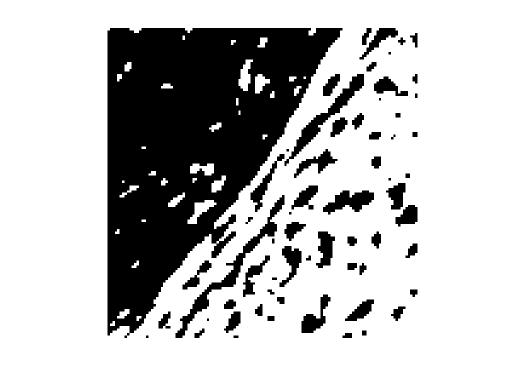}
  \caption{Generated mask after segmentation on original image: channel G}
  \label{fig:comparisonPreprocessing}
\end{figure}

\begin{figure}[htbp]
  \centering
  \includegraphics[width=0.5\linewidth]{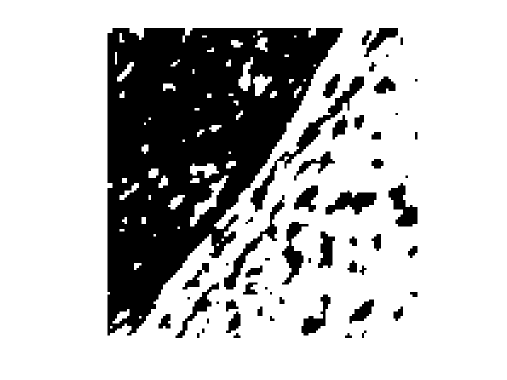}
  \caption{Generated mask after segmentation on original image: channel B}
  \label{fig:comparisonPreprocessing}
\end{figure}


\begin{figure}[htbp]
  \centering
  \includegraphics[width=0.35\linewidth]{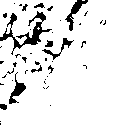}
  \caption{RGB with GN applied using Gaussian filter, generated Mask aggregated using logical OR.}
  \label{fig:comparisonPreprocessing}
\end{figure}

\begin{figure}[htbp]
  \centering
  \includegraphics[width=0.35\linewidth]{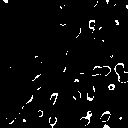}
  \caption{RGB with GN applied using Gaussian filter, generated Mask aggregated using logical AND.}
  \label{fig:comparisonPreprocessing}
\end{figure}

\begin{figure}[htbp]
  \centering
  \includegraphics[width=0.35\linewidth]{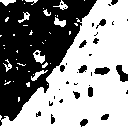}
  \caption{RGB with GN applied using Gaussian filter, generated Mask aggregated using logical Majority Vote.}
  \label{fig:comparisonPreprocessing}
\end{figure}

\begin{figure}[htbp]
  \centering
  \includegraphics[width=0.35\linewidth]{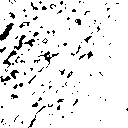}
  \caption{RGB with GN and S\&P applied using Median filter, generated Mask aggregated using logical OR.}
  \label{fig:comparisonPreprocessing}
\end{figure}

\begin{figure}[htbp]
  \centering
  \includegraphics[width=0.35\linewidth]{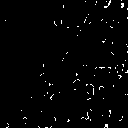}
  \caption{RGB with GN and S\&P applied using Median filter, generated Mask aggregated using logical AND.}
  \label{fig:comparisonPreprocessing}
\end{figure}

\begin{figure}[htbp]
  \centering
  \includegraphics[width=0.35\linewidth]{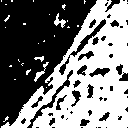}
  \caption{RGB with GN and S\&P applied using Median filter, generated Mask aggregated using logical Majority Vote.}
  \label{fig:comparisonPreprocessing}
\end{figure}

\begin{figure}[htbp]
  \centering
  \includegraphics[width=0.35\linewidth]{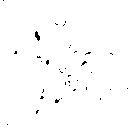}
  \caption{RGB with GN and S\&P applied using only Gaussian filter on Gaussian Noise and Salt \& Pepper, generated Mask aggregated using logical OR.}
  \label{fig:comparisonPreprocessing}
\end{figure}

\begin{figure}[htbp]
  \centering
  \includegraphics[width=0.35\linewidth]{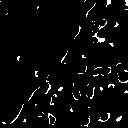}
  \caption{RGB with GN and S\&P applied using only Gaussian filter on Gaussian Noise and Salt \& Pepper, generated Mask aggregated using logical AND.}
  \label{fig:comparisonPreprocessing}
\end{figure}

\begin{figure}[htbp]
  \centering
  \includegraphics[width=0.35\linewidth]{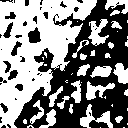}
  \caption{RGB with GN and S\&P applied  using only Gaussian filter on Gaussian Noise and Salt \& Pepper, generated Mask aggregated using logical Majority Vote.}
  \label{fig:comparisonPreprocessing}
\end{figure}

\begin{figure}[htbp]
  \centering
  \includegraphics[width=0.35\linewidth]{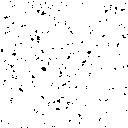}
  \caption{RGB with GN and S\&P applied using both Gaussian and Median filter, generated Mask aggregated using logical OR.}
  \label{fig:comparisonPreprocessing}
\end{figure}

\begin{figure}[htbp]
  \centering
  \includegraphics[width=0.35\linewidth]{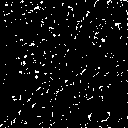}
  \caption{RGB with GN and S\&P applied using both Gaussian and Median filter, generated Mask aggregated using logical AND.}
  \label{fig:comparisonPreprocessing}
\end{figure}

\begin{figure}[htbp]
  \centering
  \includegraphics[width=0.35\linewidth]{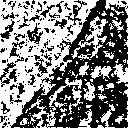}
  \caption{RGB with GN and S\&P applied using both Gaussian and Median filter, generated Mask aggregated using logical Majority Vote.}
  \label{fig:comparisonPreprocessing}
\end{figure}

Here below an example of generated mask. 


\section{Appendix A: Project results visualization}

\subsection{Preprocessing and data preparation}

As we can establish from Section 2. 'Proposed Approach', within the Preprocessing and Data Augmentation part, we can see how the image data is treated. Recall that a digital image is stored as $I \in \mathbf{R} ^ { r \times c \times n_{channels}}$, where $r$ represents the image length, whereas $c$ (\# columns), and $N_{channels}$ is the number of channels of colored images, in this case since we dealt with RGB images, it is indeed $I \in \mathbf{R}^{128  \times 128 \times 3}$.
Recall that for grey scaled images, the task is a bit easier, since the level method evolution works only on a unique channel, instead of RGB images where multiple tasks has to run for each channel, and its result is made up by the aggregation of those 3 channels. Various techniques are exploited, as shown in subsection 3.2, here below you can find experimental results on the dataset.


\begin{figure*}[p]
\centering
\includegraphics[width=\linewidth]{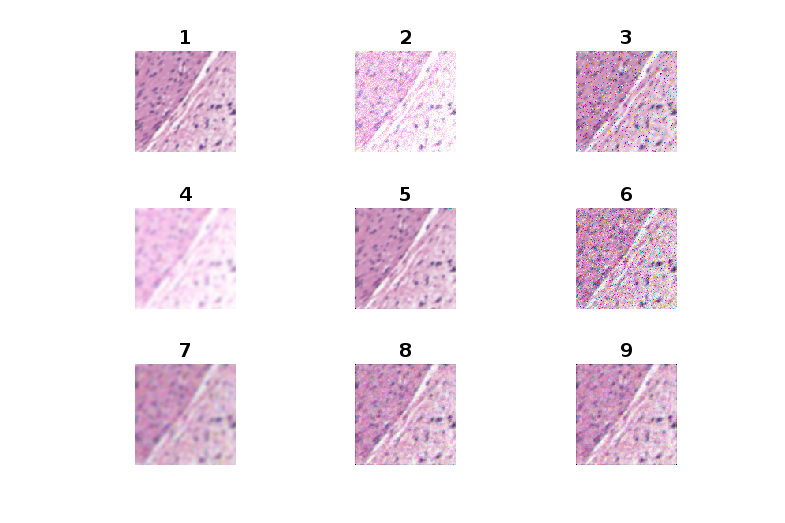}
\caption{Legend of the preprocessed images and filtered images}
\centering
\begin{tabular}{cl}
\toprule
\textbf{Label} & \textbf{Description} \\
\midrule
(1) & Original Image \\
(2) & Gaussian Noisy Image \\
(3) & Salt and Pepper Noisy Image \\
(4) & Gaussian Filtered \\
(5) & Median Filtered \\
(6) & Gaussian Salt and Pepper Image \\
(7) & Gaussian Salt and Pepper Filtered \\
(8) & Median Salt and Pepper Filtered \\
(9) & Median Gaussian Salt and Pepper Filtered \\
\bottomrule
\end{tabular}
\label{tab:legend}
\label{fig:preprocessingComparison}
\end{figure*}

\section{Appendix B: Additional run experiments examples}

\begin{figure}[htbp]
  \centering
  \includegraphics[width=0.65\linewidth]{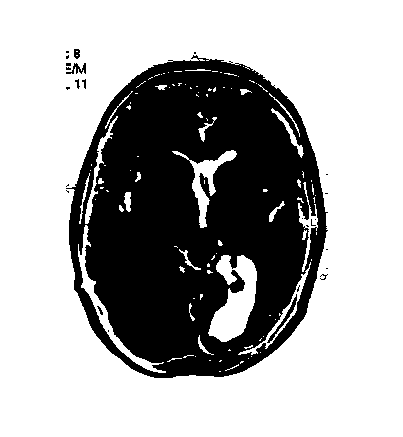}
  \caption{Generated segmented mask MRI brain with tumor}
  \label{fig:mri_brain_mask}
\end{figure}

Here below in figure \ref{fig:mri_segmented} we can also see an additional result run using MRI (Magnetic resonance imaging) as initial input as shown in figure \ref{fig:mri_brain_mask}.

\begin{figure}[htbp]
  \centering
  \includegraphics[width=0.65\linewidth]{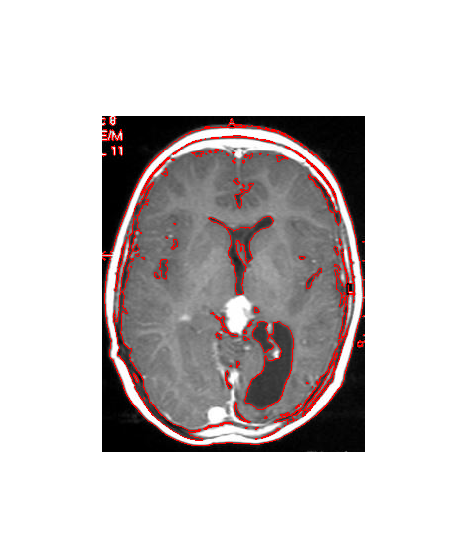}
  \caption{Segmented image MRI brain with tumor}
  \label{fig:mri_segmented}
\end{figure}

As we can see from the segmented mask, the area in black filled with unsual white patterns shows us a tumor alike pattern. 

\section{Discussion}
In this paper we saw how the Chan-Vese method can be applied to a wide range of images accomplishing overall good results, even with images subject to various types of noise. The project could be expanded by exploiting the loss implemented both in \texttt{MATLAB} and in \texttt{python} is possible to train on a custom dataset to further explore better optimization techniques for increased performances in the loss module, in contrast to new SoAT methodologies.


\begin{thebibliography}{99}
\bibitem{mumford1989optimal}
D. Mumford and J. Shah, ``Optimal approximations by piecewise smooth functions and associated variational problems,'' \emph{Communications on Pure and Applied Mathematics}, vol. 42, no. 4, pp. 577--685, 1989.

\bibitem{chan2001active}
T. F. Chan and L. A. Vese, ``Active contours without edges,'' \emph{IEEE Transactions on Image Processing}, vol. 10, no. 2, pp. 266--277, 2001.

\bibitem{aubert2003image}
G. Aubert, M. Barlaud, O. Faugeras, and S. Jehan-Besson, ``Image segmentation using active contours: calculus of variations or shape gradients?,'' \emph{SIAM Journal on Applied Mathematics}, vol. 63, no. 6, pp. 2128--2154, 2003.

\bibitem{boykov2004experimental}
Y. Boykov and V. Kolmogorov, ``An experimental comparison of min-cut/max-flow algorithms for energy minimization in vision,'' \emph{IEEE Transactions on Pattern Analysis and Machine Intelligence}, vol. 26, no. 9, pp. 1124--1137, 2004.

\bibitem{bregman1967relaxation}
L. Bregman, ``The relaxation method of finding the common points of convex sets and its application to the solution of problems in convex programming,'' \emph{USSR Computational Mathematics and Mathematical Physics}, vol. 7, pp. 200--217, 1967.
\end{thebibliography}
\end{document}